\documentclass[letterpaper, 10 pt, conference]{ieeeconf}  

\IEEEoverridecommandlockouts                              

\usepackage{xcolor}
\definecolor{own_purple}{HTML}{A043F9}

\usepackage{hyperref}
\usepackage{todonotes}
\usepackage[caption=false]{subfig}
\usepackage{multirow}

\usepackage{tikz}
\usetikzlibrary{positioning,shapes.geometric,automata,decorations.pathreplacing,shapes.arrows,arrows,shapes,arrows.meta}

\newcommand\copyrighttext{%
\footnotesize \copyright 2022 IEEE. Personal use of this material is permitted.  Permission from IEEE must be obtained for all other uses, in any current or future media, including reprinting/republishing this material for advertising or promotional purposes, creating new collective works, for resale or redistribution to servers or lists, or reuse of any copyrighted component of this work in other works.\\
\url{https://doi.org/10.1109/ICRA46639.2022.9812417}
}
\newcommand\copyrightnotice{%
	\begin{tikzpicture}[remember picture,overlay]
		\node[anchor=south,yshift=0pt] at (current page.south) {\fbox{\parbox{\dimexpr\textwidth-\fboxsep-\fboxrule\relax}{\copyrighttext}}};
	\end{tikzpicture}%
}

\title{\LARGE \bf
\textsc{ROS2swarm} - A ROS~2 Package for Swarm Robot Behaviors}

\author{Tanja Katharina Kaiser,$^{1}$  Marian Johannes Begemann,$^{1}$ Tavia Plattenteich,$^{1}$ \\ Lars Schilling,$^{2}$ Georg Schildbach,$^{3}$ and Heiko Hamann$^{1}$
\thanks{$^{1}$TKK, MJB, TP, HH are with the Institute of Computer Engineering, University of L\"ubeck, Germany {\tt\small kaiser@iti.uni-luebeck.de}}%
\thanks{$^{2}$LS is with the Institute of Robotics, University of L\"ubeck, Germany}%
\thanks{$^{3}$GS is with the Institute for Electrical Engineering in Medicine,\linebreak University of L\"ubeck, Germany}%
}

\begin{document}

\maketitle
\copyrightnotice
\thispagestyle{empty}
\pagestyle{empty}

\begin{abstract}

Developing reusable software for mobile robots is still challenging. Even more so for swarm robots, despite the desired simplicity of the robot controllers. Prototyping and experimenting are difficult due to the multi-robot setting and often require robot-robot communication. 
Also, the diversity of swarm robot hardware platforms increases the need for hardware-independent software concepts.
The main advantages of the commonly used robot software architecture ROS~2 are modularity and platform independence.
We propose a new ROS~2~package, \textsc{ROS2swarm}, for applications of swarm robotics that provides a library of ready-to-use swarm behavioral primitives. 
We show the successful application of our approach on three different platforms, the TurtleBot3 Burger, the TurtleBot3 Waffle Pi, and the Jackal UGV, and with a set of different behavioral  primitives, such as aggregation, dispersion, and collective decision-making. 
The proposed approach is easy to maintain, extendable, and has good potential for simplifying swarm robotics experiments in future applications.
\end{abstract}

\section{Introduction}

Robot swarms~\cite{Hamann2018} are decentralized systems in which relatively simple robots solve tasks collectively. 
One of their great advantages is the robustness of the system; there is no single point of failure as the defect of some swarm members will not prevent successful task execution. 

A~common software engineering challenge in mobile robotics is the development of reusable robotic software, which 
``is difficult primarily due to the variability in robotic platforms''~\cite{nesnas2006claraty}. Software engineering challenges specific to swarm robotics are ``the lack of general tools for experimentation'', ``common `swarm libraries' do not yet exist, and reusing code is difficult owing to the lack of swarm-centric development platforms"~\cite{pinciroli2016buzz}.
The arguably most prominent effort in mobile robotics to push for common software architectures and reusable code is the Robot Operating System (ROS)~\cite{quigley2009ros}.
While ROS~\cite{quigley2009ros} is widely used in both single mobile robot applications and multi-robot systems, it is rarely used for swarm robotics to date~\cite{dorigo21}. 
This is, among other reasons, due to the central ROS master node which counteracts the paradigm of decentralization in swarm robotics. 
However, ROS~2~\cite{Maruyama2016} replaces the central ROS master node with a Data Distribution Service (DDS) as a middleware and is thus decentralized. 
This change makes ROS~2 interesting for swarm robotics research as it brings about several advantages, such as easy reusability of code on different robot platforms.  
Several works on distributed multi-robot systems and robot swarms using ROS~2 were recently published, ranging from robots that synchronize and swarm~\cite{barcis2019,barcis2020sandsbots} over a toolbox for distributed control schemes for heterogeneous multi-robot systems~\cite{Testa2021} to concepts for large-scale scalable micro-aerial vehicle swarms~\cite{queralta2021towards}.

\begin{figure}
    \centering
     \subfloat[Five TurtleBot3 Waffle Pis]{\includegraphics[width=0.48\linewidth]{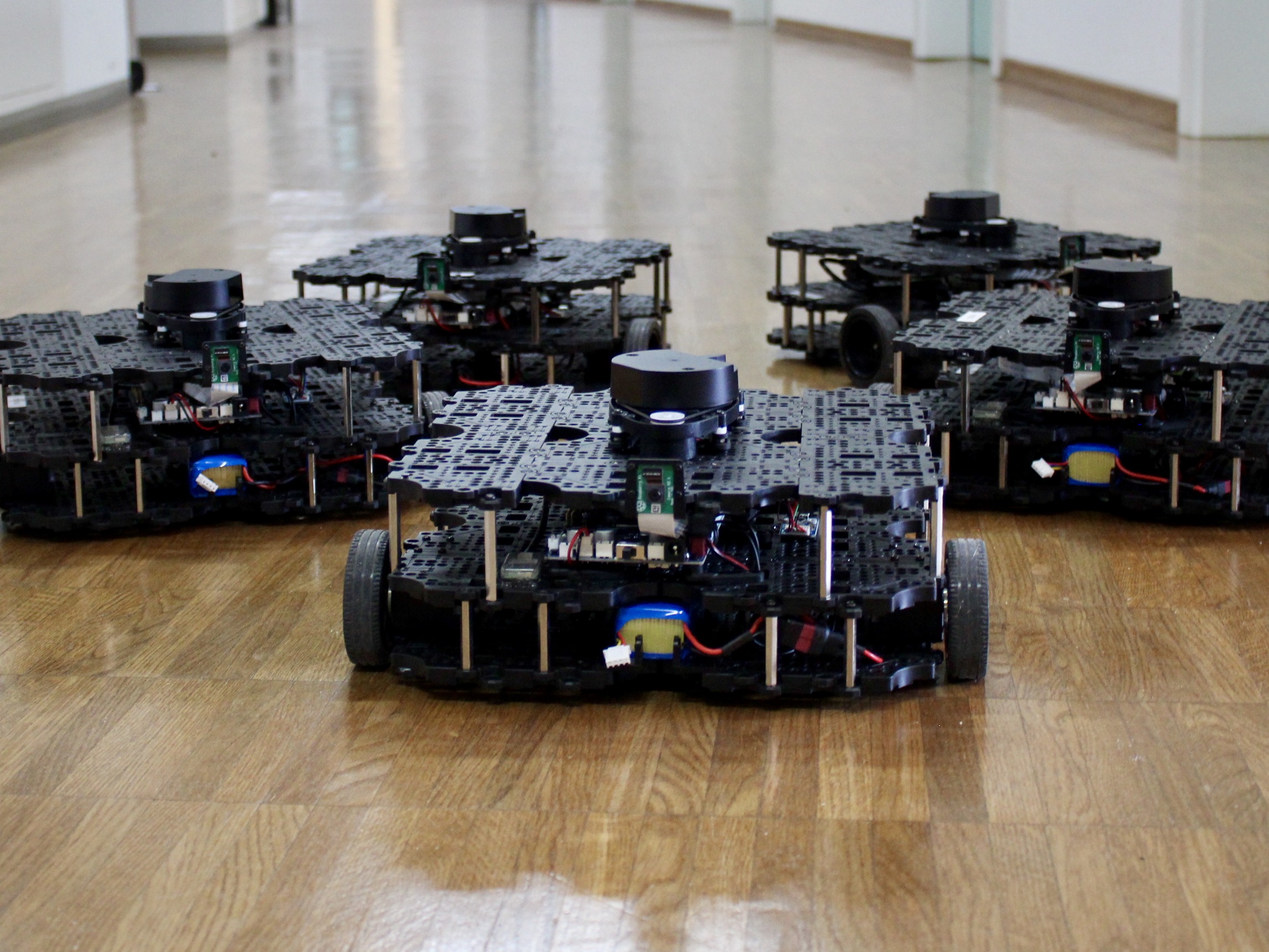}\label{fig:tb3}}
     \hspace{2mm}
     \subfloat[Jackal UGV]{\includegraphics[width=0.48\linewidth]{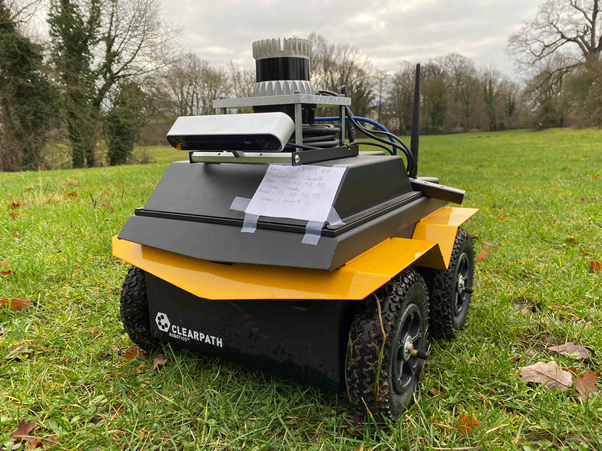}\label{fig:jackal}}
    \caption{Two of the three used robot hardware platforms: (a)~a~swarm of the ground mobile robot platform TurtleBot3 Waffle Pi from Robotis and (b)~a~Jackal unmanned ground vehicle (UGV) from Clearpath Robotics.}
\label{fig:robots}
\end{figure}

In swarm robotics systems, both hardware and control are usually kept simple relying on self-organization and emergence~\cite{Hamann2018}.
Despite the simple reactive controllers of individual robots, complex behaviors emerge due to the collaboration between robots. 
There are several primitive behaviors in swarm robotics, that are frequently used and studied, such as aggregation (i.e., robots group in one spot), dispersion (i.e., robots distribute while staying in contact), and flocking (i.e., bird-inspired coordinated motion). 

As ROS~2 allows for the reuse of code on various robot platforms, a ROS~2 package providing such swarm behavior primitives allows for an easy and rapid setup of swarm robot experiments. 
We introduce \textsc{ROS2swarm}, a ROS~2 package that provides an easy-to-extend framework for and a library of swarm robot behaviors for mobile robots. 
Included swarm behaviors can be used out of the box with any ROS or ROS~2 compatible mobile robot platform that provides sensor data via a scan message and sets the linear speed along the x-axis and the rotational speed around the z-axis. 
Since we are building on the modularity of ROS~2, new behaviors can easily be added and used as modules to build more complex behaviors. 

\enlargethispage{1\baselineskip}
In the next section, we discuss design patterns for collective behaviors as well as existing ROS and ROS~2 packages for swarm robotics. 
Section~\ref{sec:ros2swarm} presents the structure and implementation of \textsc{ROS2swarm} and introduces the robot platforms that are already supported. 
In Sec.~\ref{sec:experiments}, we show the functionality and benefits of our \textsc{ROS2swarm} package in simulation and real robot experiments using the TurtleBot3 Burger, the TurtleBot3 Waffle Pi (see Fig.~\ref{fig:tb3}), and the Jackal UGV~(see Fig.~\ref{fig:jackal}) robots.  
Finally, we summarize our work and conclude with an outlook on on-going and future work that will be included in the package in Sec.~\ref{sec:conclusion}.

\section{Related Work} \label{sec:relatedwork}

To our knowledge, we are the first to propose a ROS~2 package that implements modular and reusable design patterns for swarm behaviors using a decentralized setup in simulations and on real robot hardware.
We were inspired by several works on design patterns to facilitate the engineering of distributed multi-agent systems when creating the modular framework.  
De Wolf and Holvoet~\cite{deWolf2006}, for example, consolidate best practices for engineering self-organizing emergent systems in a catalog of design patterns for decentralized coordination mechanisms, such as digital pheromones. 
Fernandez-Marquez et al.~\cite{fernandez-marquez_description_2013} go one step further by proposing modular and reusable design patterns for bio-inspired mechanisms for self-organizing systems. 
They organize patterns into three layers, including basic mechanisms and two levels of composed patterns. 
Pitonakova et al.~\cite{pitonakova2018} focus specifically on robot swarms by presenting information exchange design patterns for the example of foraging while other swarm tasks are left subject to future work. 
Loosely inspired by these works, we create a structure of basic and combined patterns that allows for the implementation of modular and reusable swarm behaviors in our \textsc{ROS2swarm} package. 

Already existing ROS and ROS~2 packages for swarm robotics applications do not create such a modular and reusable framework, or they even rely on central components. 
Most of the existing packages rely on ROS and its central ROS master node. 
\textit{ROSBuzz}~\cite{stonge2020}, for example, integrates the swarm-oriented programming language Buzz~\cite{pinciroli2016buzz} with ROS to ease the deployment of heterogeneous robot swarms in the field. 
However, a global positioning system is required.
Similarly, the ROS programming framework \textit{micros\_swarm\_framework}~\cite{micros_swarm_framework} is inspired by the Buzz programming language and provides abstractions and swarm functions to facilitate the implementation of interaction and communication between swarm members in swarm applications. 
\textit{CMUSWARM}~\cite{morris2018full} is a full stack swarm architecture in ROS that allows for the design, deployment, and evaluation of swarm algorithms. 
But the framework relies on a centralized core layer for the deployment of the algorithms to the swarm. 
Hrabia et al.~\cite{hrabia2018} present a modular and reusable ROS framework for self-organization in multi-robot systems. 
Self-organization mechanisms are integrated based on an adapted version of the bio-inspired design patterns by Fernandez-Marquez et al.~\cite{fernandez-marquez_description_2013}. 
Other swarm related ROS packages focus on providing swarm simulations. 
\textit{swarm\_stage\_ros}~\cite{KalempaTeixeiraOliveiraFabro2018}, for example, configures a swarm robot scenario for a cleaning task in the Stage simulator automatically, and \textit{swarm\_robot\_ros\_sim}~\cite{swarm_robot_ros_sim} provides a generic swarm robot simulation platform for decentralized formation control. 
Unlike all the previously described ROS packages, \textit{\textsc{ChoiRbot}}~\cite{Testa2021} is based on ROS~2. It provides a toolbox for cooperative robots to facilitate the implementation of optimization-based distributed control schemes, but requires global position information of all robots. 
In contrast, our approach of \textsc{ROS2swarm} is decentralized as it relies on ROS~2, does not require any global information, such as robot positions, and allows for the easy implementation and execution of swarm behaviors in simulations and on real robots.

\section{\textsc{ROS2swarm} package} \label{sec:ros2swarm}

\textsc{ROS2swarm}\footnote{https://gitlab.iti.uni-luebeck.de/ROS2/ros2swarm. Currently available for ROS~2 Dashing Diademata.} provides a library of ready-to-use behavioral primitives for swarm robotics applications that can easily be extended. 
Following the key concept of distributed control in robot swarms, the package and consequently the swarm behaviors are executed on each robot autonomously and independently. 
For some behaviors, data sharing between robots is needed and implemented using global namespace ROS~2 topics requiring all swarm members to be in the same network.  
In the following, we introduce the general package structure of \textsc{ROS2swarm}, the pattern hierarchy for the implementation of swarm behaviors, and how the package can easily be used with different ground mobile robots in the Gazebo simulator~\cite{gazebo_webpage} and on real robots.

\subsection{General Package Structure} 

\begin{figure}
    \centering
\includegraphics[width=1.0\linewidth]{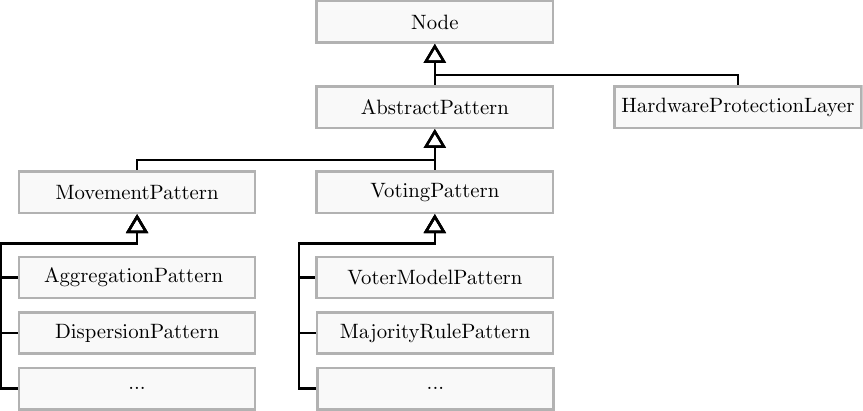}
    \caption{UML diagram of \textsc{ROS2swarm}'s package architecture. The \textit{AbstractPattern} class inherits from the standard ROS~2 \textit{node} class. \textit{MovementPattern} and \textit{VotingPattern} are subclasses of the \textit{AbstractPattern} class and differentiate implemented behaviors into patterns for robot movement and patterns for collective decision-making. Left out of the pattern hierarchy is the \textit{HardwareProtectionLayer} that prevents collisions with obstacles.
    Not shown are several utility classes for processing laser scan data, better readability of state machines, and handling of common voting procedures and standard ROS~2 processes. }
    \label{fig:UMLMP}
\end{figure}

\textsc{ROS2swarm} consists of a main ROS~2 Python package providing the swarm behaviors and an accompanying C++ package with custom ROS~2 message interfaces. 
The main package is structured by a class hierarchy as shown in Fig.~\ref{fig:UMLMP}. 
The \textit{AbstractPattern} class serves as the basis for the implementation of all swarm behaviors. 
We distinguish between two different types of patterns: \textit{movement patterns}, which control the motion of the robot swarm, and \textit{voting patterns}, which implement collective decision-making. 
For the voting patterns, additional ROS~2 message interfaces that allow exchanging opinions between robots are provided by the accompanying C++ package. 
Both movement patterns and voting patterns can be implemented as standalone \textit{basic patterns} or by combining multiple patterns into a \textit{combined pattern}. 
Termination conditions for each pattern can easily be included, for example, stopping the execution of a pattern until timeout.
Hence, we obtain a modular package structure that allows for reusing patterns. 

\begin{figure}
    \centering
\includegraphics[width=0.9\linewidth]{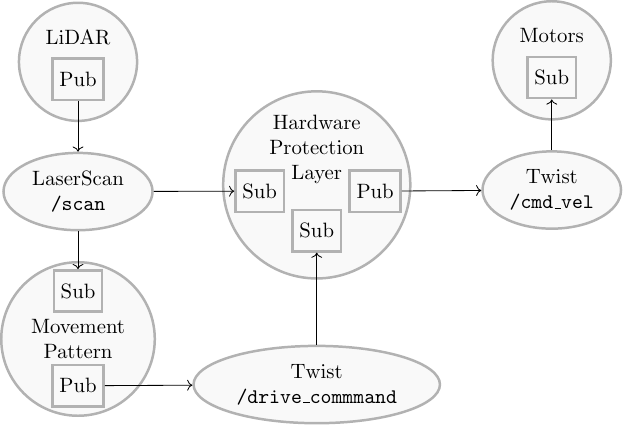}
    \caption{Communication between the ROS~2 nodes of \textsc{ROS2swarm} using the ROS~2 publisher/subscriber model when executing movement patterns.
    Laser scan data from a LiDAR is used by the movement pattern to calculate a drive command and by the hardware protection layer to check for potential collisions. 
    The hardware protection layer sends either the drive command from the movement pattern or an adjusted command to avoid obstacles to the robot's motors. 
    ROS~2 nodes are represented by circles and ROS~2 topics by ellipses. Publishers (\textit{Pub}) and subscribers (\textit{Sub}) are represented by rectangles.  Arrows indicate the data flow.
    } \label{fig:MParchitecture}
\end{figure}

The parameters of each pattern are set via YAML files so that the behaviors can easily be adjusted for individual applications. 
We include a set of these parameter files for each robot platform (i.e., TurtleBot3 Waffle Pi, TurtleBot3 Burger, and Jackal so far) to enable easy switching between different platforms. 
Patterns are executed independently for each robot and can easily be started via launch scripts that handle the start of all required ROS~2 nodes with the specified parameterization. 
In addition, the launch scripts remap ROS~2 nodes to a robot's individual namespace,  where it is required to ensure correct data handling in the swarm. 
Topics in the global namespace are only used for essential neighbor-to-neighbor communication within the swarm, for example, as required by voting patterns. 
In the current setup of robot experiments with \textsc{ROS2swarm}, a robot has no information about the distance to a swarm member that shared a message via the global namespace topic. 
Thus, all other swarm members are considered as a robot's neighbors to date. 
Our architecture allows for easily adding new patterns by creating a new subclass of either the \textit{MovementPattern} class or the \textit{VotingPattern} class that implements the pattern logic as well as a launch script and a parameter file per robot platform. 

We include a \textit{Hardware Protection Layer} next to the pattern hierarchy to prevent collisions with obstacles, including other robots, and possible hardware damage to robots, 
that runs permanently and independently from the executed swarm behavior.
As it is especially relevant when executing movement patterns, we explain the hardware protection layer in more detail in Sec.~\ref{sec:movement}.
The hardware protection layer is left out of the pattern hierarchy to ensure its stability independent from the implemented swarm behaviors. 
Furthermore, several utility classes provide common methods to process laser scan messages and to handle voting procedures, and an extendable Enum with a basic set of states for better readability of implemented state machines. 
We verified our code by running several tests. 
Next to the ROS~2 packages constituting \textsc{ROS2swarm}, we provide several ROS and ROS~2 packages and launch scripts for running the implemented behavioral primitives on simulated swarms in Gazebo and on real robots.

\subsection{Movement Patterns} \label{sec:movement}

\begin{figure}
    \centering
\includegraphics[width=0.9\linewidth]{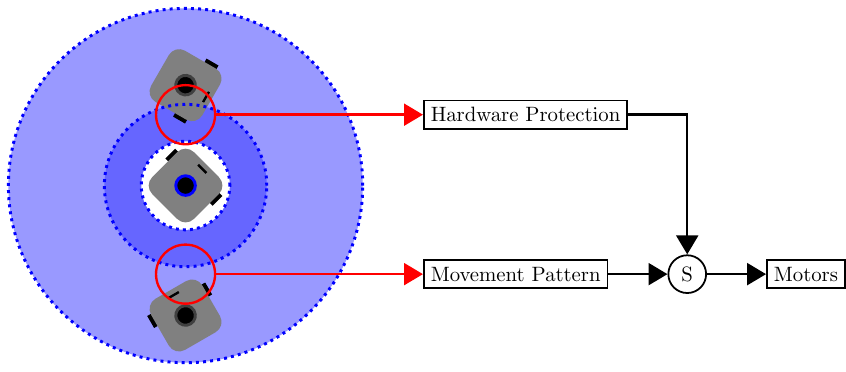}
    \caption{Arbitration architecture of the hardware protection layer. 
    The light blue area indicates the sensor range which is used by the movement pattern to calculate drive commands. 
    Hardware protection is active and suppresses (using `suppression'~\textit{S} as defined in the subsumption architecture) drive commands from the movement patterns when there are obstacles in the dark blue area as shown for the upper robot.} \label{fig:HWP}
\end{figure}

Movement patterns implement behavioral primitives that initiate and guide the robot motion.
Six movement patterns are included in the current version of \textsc{ROS2swarm}: \textit{attraction}, \textit{dispersion}, \textit{discussed dispersion}, \textit{drive}, \textit{random walk},  and \textit{flocking}~(cf. Table~\ref{tab:algorithms}). 

\begin{table*}
\caption{Behaviors that are currently included as patterns in the \textsc{ROS2swarm} package.\label{tab:algorithms}}
\centering
\begin{tabular}{lll}
\hline
 \textbf{} & \textbf{pattern} & \textbf{description} \\ \hline 
\multirow{6}{*}{\rotatebox[origin=c]{90}{\textbf{movement}}} & attraction & Aggregation of robots based on an attractive potential field. \\
 & dispersion & Distribution of robots based on a repulsive potential field.\\
& discussed dispersion & Distribution of robots while maintaining a distance decided on by the swarm. \\
& drive & Driving straight ahead. \\ 
& flocking  & Minimalist flocking algorithm based on Moeslinger et al.~\cite{moeslinger2011}. \\ 
& random walk  & Random walk based on a simple state machine switching between driving straight ahead and turning randomly. \\ \hline    
\multirow{3}{*}{\rotatebox[origin=c]{90}{\textbf{voting}}} & & \\[-1ex]
 & majority rule &  Opinion update based on the majority opinion. \\ 
& voter model & Opinion update based on adopting the opinion of a random neighbor.\vspace{1.3ex} \\
\hline 
\end{tabular}
\end{table*}

Fig.~\ref{fig:MParchitecture} visualizes the interplay between the different ROS~2 nodes that are used for movement patterns via the ROS~2 publisher/subscriber model. 
Each robot executes the movement pattern independently from the rest of the swarm. 
The movement pattern receives the robot's laser scan data (e.g., from a LiDAR), processes it, and calculates a drive command (i.e., linear speed and rotation speed). 
The hardware protection layer receives both the laser scan data and the drive command from the movement pattern. 
It uses an arbitration architecture to decide based on the distance to the robot's nearest obstacle whether to send the drive command determined by the movement pattern or a drive command for collision avoidance to the robot's motors, see Fig.~\ref{fig:HWP}. 
The distance threshold to trigger hardware protection can be set via a parameter and can thus be adjusted for different mobile robot platforms.
If there is at least one obstacle, including other robots, within the specified range, hardware protection sends a drive command to the robot calculated based on a repulsive potential field approach. 
This way robots should keep a distance to obstacles, but we cannot guarantee to prevent all collisions.
Otherwise, the drive command determined by the movement pattern is executed. 
Hardware protection determines whether obstacle avoidance is necessary every time new laser scan data is received, regardless of whether also a new drive command has been received from the movement pattern.  
Therefore, movement patterns have to publish drive commands regularly to ensure their correct execution.

\subsection{Voting Patterns}

Voting patterns implement collective decision-making behaviors. Collective decision-making is an important capability of a robot swarm to implement consistent actions on swarm level~\cite{valentini16a,valentini2017best}.
Currently, \textsc{ROS2swarm} includes two voting patterns: \textit{majority rule} and \textit{voter model} (cf. Table~\ref{tab:algorithms}). 

Each voting pattern uses a common ROS~2 topic in the global namespace to enable swarm members to exchange their opinions via a custom ROS~2 message type that includes the robot's ID and its opinion represented by an Integer. 
All swarm members have to be in the same network in order
to have access to a shared global namespace topic.  
Based on the other robots' opinions, robots then update their own opinion using the decision rule implemented by the pattern. 
The currently included voting patterns make use of a tumbling time window approach, that is, the data stream is split into fixed-size, contiguous time intervals that do not overlap.
Other options, such as a sliding time window approach (i.e., time intervals can contain overlapping data) or including synchronization, are possible within the framework and can be implemented if required. 
The data received during one time frame is then processed by the decision rule to find a new opinion.
Robots share their opinions after each update and for the most part, the swarm eventually reaches consensus. 
By contrast to classical swarm settings that rely purely on local communication, each swarm member can currently access the opinions of the full swarm via the global namespace topic.

\subsection{Robot Platforms} \label{sec:robots}

\textsc{ROS2swarm} is designed to work with any ground mobile robot that (i)~supports ROS or ROS~2, (ii)~has sensors that cover the robot's surroundings and publish laser scan data, and (iii)~can be controlled by specifying a linear and a rotational speed.  
Although \textsc{ROS2swarm} is implemented in ROS~2, it sill can be used with robot platforms running on ROS by using a network bridge~\cite{Thomas2015} that enables the exchange of messages between ROS and ROS~2.
To include support for a new robot platform, a configuration file setting the robot's parameters, launch scripts for the robot platform, and parameter files for the patterns have to be added to the existing package. 
We provide out-of-the-box functionality in the Gazebo simulator and for real robots for three ground mobile robot platforms: Robotis TurtleBot3 Waffle Pi, Robotis TurtleBot3 Burger, and Clearpath Robotics Jackal UGV. 

The TurtleBot3 mobile robots are small and modular ROS and ROS~2 research and education platforms for indoor environments that are controlled via a Raspberry Pi. 
The TurtleBot3 Burger and the TurtleBot3 Waffle Pi (cf. Fig.~\ref{fig:tb3}) differ mainly in size, that is, the TurtleBot3 Burger has a smaller footprint and is taller than the TurtleBot3 Waffle Pi. 
Both have differential drive, an IMU, and a LiDAR with $360^\circ$ field of view and a range of $0.12~\textrm{m}$ to $3.5~\textrm{m}$. 
Additionally, the TurtleBot3 Waffle Pi has a Raspberry Pi camera. 

The Jackal robots (cf. Fig.~\ref{fig:jackal}) are mobile research platforms for outdoor environments. They are equipped with an onboard computer, GPS, and IMU by default. Our version is additionally equipped with a Jetson TX2 board, a ZED stereo camera, and an Ouster OS1-16 LiDAR with 16 vertical layers, $360^\circ$ field of view, and a range of $0.8~\textrm{m}$ to $5~\textrm{m}$.
In contrast to the TurtleBot3 mobile robots, the Jackal supports only ROS to date. 
Nevertheless, \textsc{ROS2swarm} can be used with the robot platform by using the aforementioned network bridge~\cite{Thomas2015} that enables the exchange of messages between ROS and ROS~2.

\section{Experimental Evaluation} \label{sec:experiments} 

We show the versatility of our \textsc{ROS2swarm} package by running several experiments using TurtleBot3 Burger, TurtleBot3 Waffle Pi, and Jackal robots (cf. Sec.~\ref{sec:robots}). 
In Experiment~1, we execute the basic attraction pattern on the three different robot platforms in simulation and on the real TurtleBot3 Waffle Pi to illustrate how swarm behaviors can be used out of the box using \textsc{ROS2swarm}. 
In Experiment~2, we combine basic patterns into a more complex pattern to showcase the modularity of our package. 
We limit ourselves to these two experiments here due to space constraints, but additional swarm behaviors (cf. Table~\ref{tab:algorithms}) are shown in the supplementary video.

\subsection{Experiment 1: Attraction}

The patterns included in the swarm behavior library of \textsc{ROS2swarm} can be executed out of the box on multiple platforms. 
Here, we show the execution of the basic \textit{attraction} movement pattern on TurtleBot3 Burger, TurtleBot3 Waffle Pi, and Jackal robots in the Gazebo simulator and in hardware on real TurtleBot3 Waffle Pi robots. 
The pattern uses an attractive potential field approach to determine drive commands that guide the robots towards all detected obstacles, including other robots, 
within a sensor range (hereafter referred to as \textit{attraction range}) that is specified via the pattern's parameter file (see Table~\ref{tab:algorithms}). 
The LiDAR does not allow to differentiate robots and other obstacles and thus, robots can be attracted to walls or other obstacles instead of to other swarm members (sensors to discriminate walls from robots could be added). 
The hardware protection layer prevents collisions during pattern execution by ensuring that robots maintain a minimum distance to all obstacles, even when they attempt to group. 
In all experiments, we set the distance threshold for hardware protection to $0.5~\textrm{m}$ for TurtleBot3 robots and $1.2~\textrm{m}$ for Jackal robots taking the minimum detection ranges of the LiDARs into account.

\begin{figure}
    \subfloat[Initial robot positions \mbox{exemplified} by the TurtleBot3 Waffle Pi \mbox{experiment}]{\includegraphics[width=0.47\linewidth]{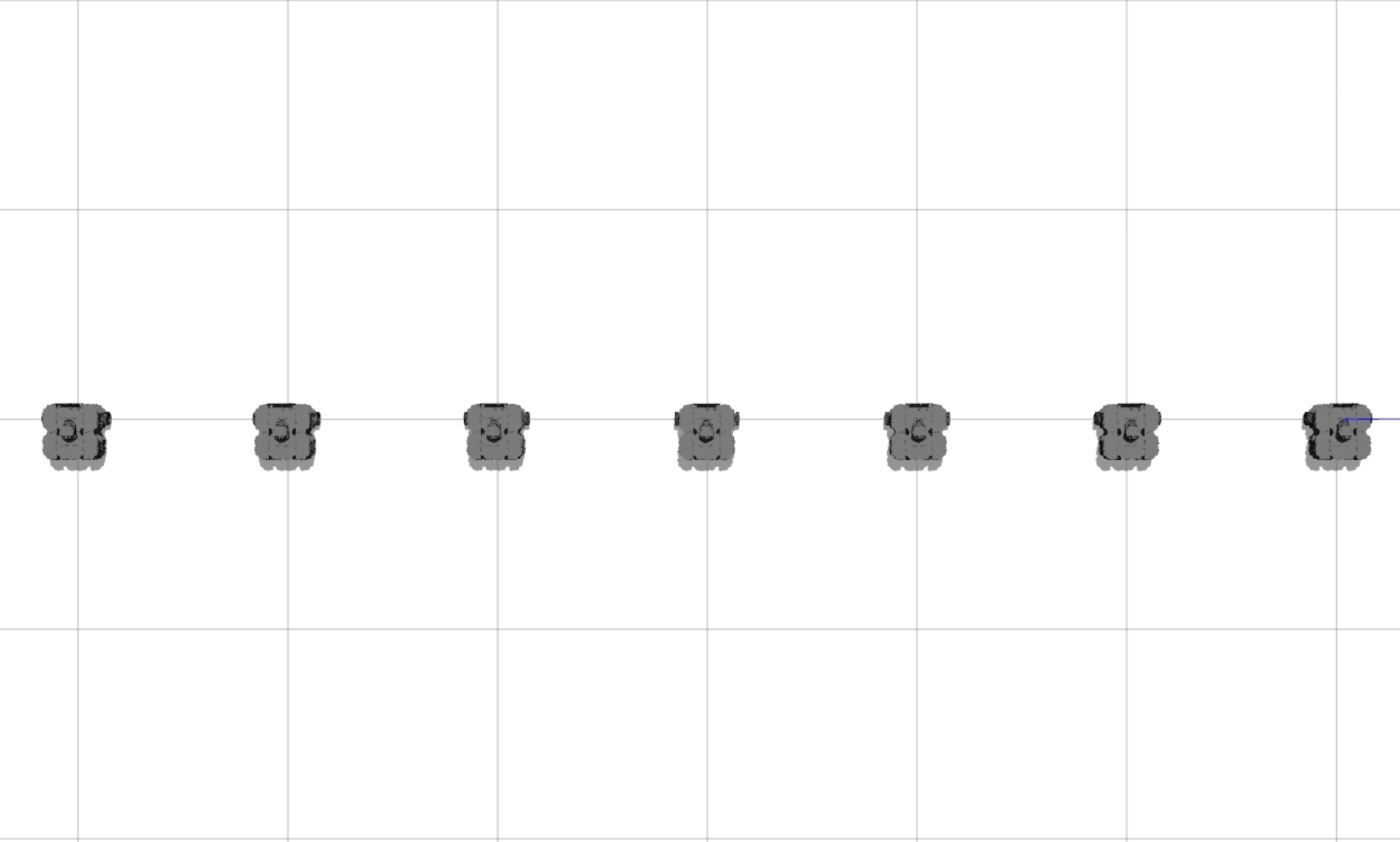}
    }\hspace{1mm}
    \subfloat[Final configuration for the grouping of seven TurtleBot3 \mbox{Burgers} \label{fig:attraction_burger}
    ]{\includegraphics[width=0.47\linewidth]{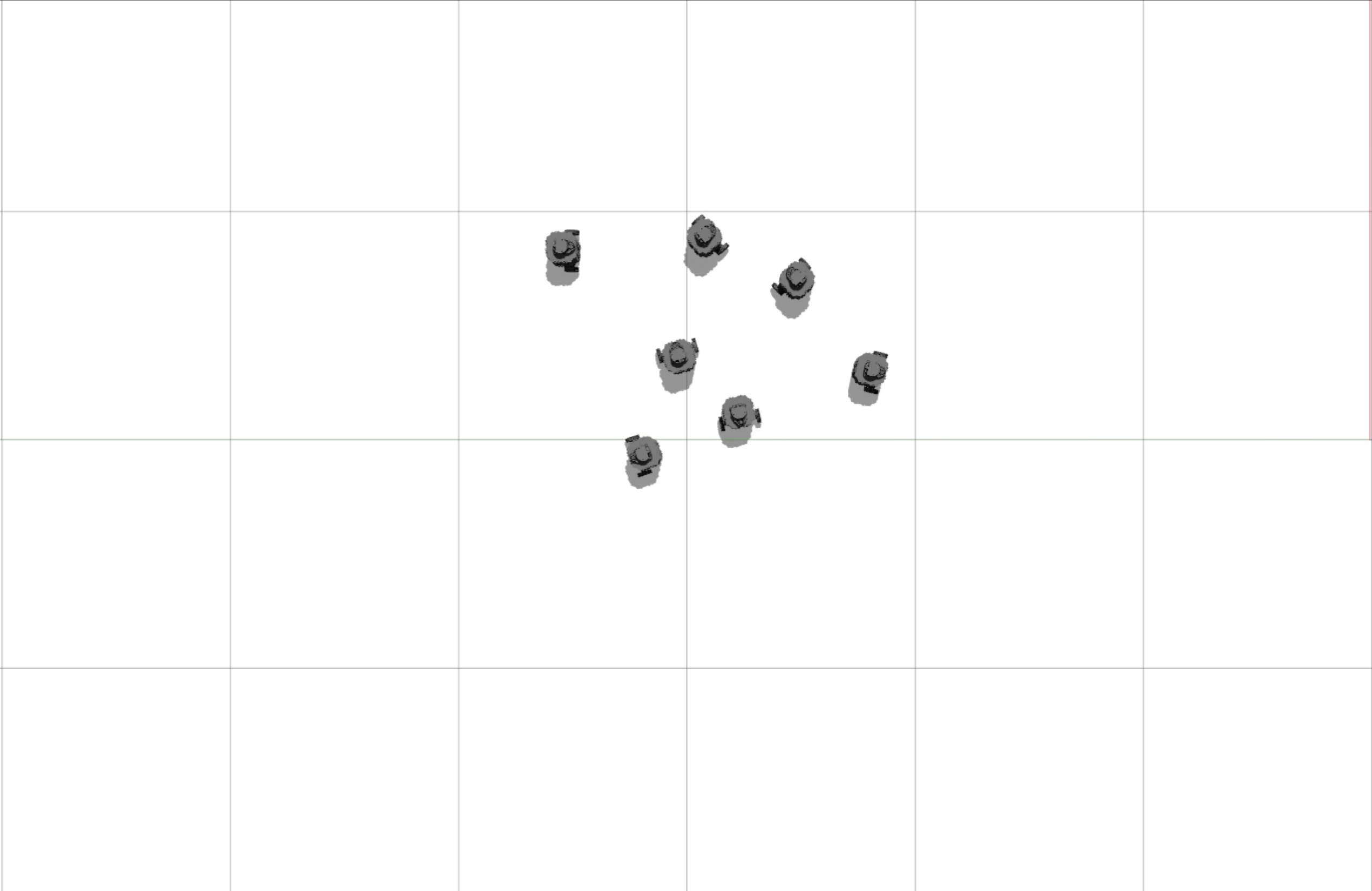}} \\
    \subfloat[Final configuration for the grouping of seven~TurtleBot3 \mbox{Waffle} Pis \label{fig:attraction_waffle}]{\includegraphics[width=0.47\linewidth]{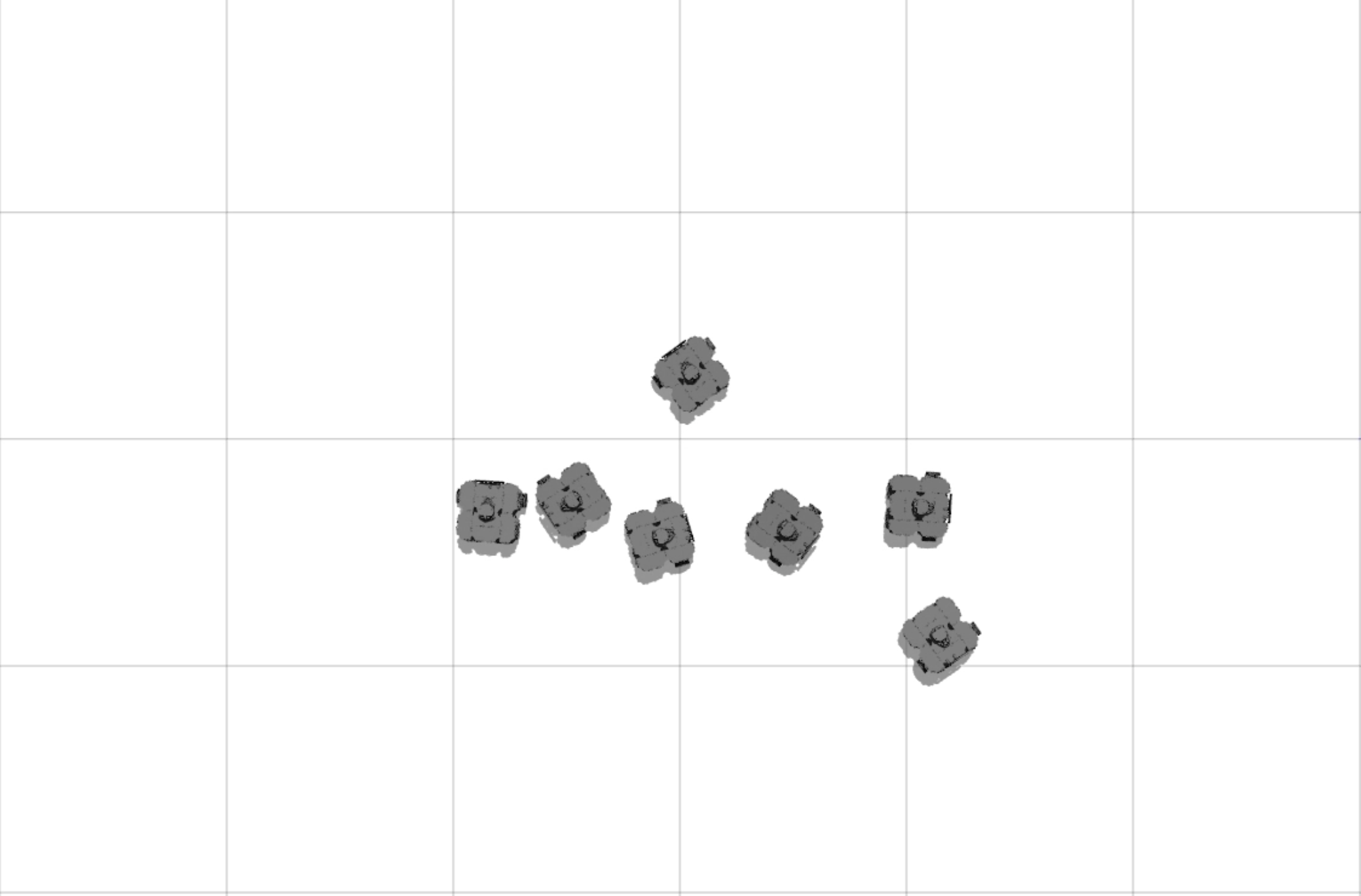}}\hspace{1mm} 
     \subfloat[Final configuration for the grouping of five Jackals \label{fig:attraction_jackal}]{\includegraphics[width=0.47\linewidth]{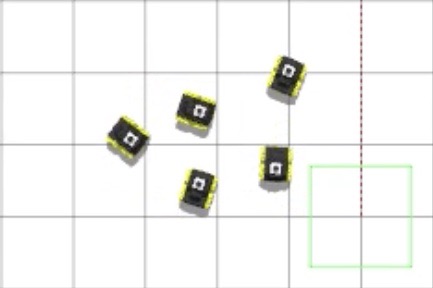}}
\caption{Experiment 1 in the Gazebo simulator: swarms of (a, c)~seven TurtleBot3 Waffle Pis, (b)~seven TurtleBot3 Burgers, and (d)~five Jackal robots executing the attraction pattern. 
Robots are initially positioned at one meter intervals whereby the rightmost robot is positioned in the arena's center. The arena has a size of $18~\textrm{m}\times 18~\textrm{m}$, but we only show a subarea for better visualization. 
Grid cells have a size of $1~\textrm{m}\times 1~\textrm{m}$ each. 
\label{fig:sim_exp}
}
\end{figure}

First, we run the attraction pattern on three different  swarms in the Gazebo simulator, namely a swarm of seven~TurtleBot3 Burgers, a swarm of seven~TurtleBot3 Waffle Pis, and a swarm of five Jackals.
A launch script provided by \textsc{ROS2swarm} handles the startup of the simulation environment. 
The script creates a swarm of the respective robot platform and executes the pattern as specified by its parameters.  
We use an empty $18~\textrm{m}\times 18~\textrm{m}$ arena that is bounded by walls and position all robots initially close to the arena center, such that the walls are out of the LiDAR's detection range. 
As the different robot platforms differ in dimensions and components, we have to parameterize the attraction pattern individually for each platform. 
We set the attraction range from the minimum detection range of the LiDAR, that is, $0.12~\textrm{m}$ for both TurtleBot3 robot platforms and $0.8~\textrm{m}$ for Jackal robots, to a maximum of $2~\textrm{m}$ for the TurtleBot3s and of $3~\textrm{m}$ for  Jackals.
In Fig.~\ref{fig:sim_exp} we show screenshots of Gazebo simulation runs of the attraction pattern on robot swarms of the three different robot platforms. 
In all simulations, the swarms successfully gather in one cluster. 
But the varying distance threshold for hardware protection, which specifies the minimum distance robots try to keep to prevent collisions, leads to differences in the grouping behavior. 
The TurtleBot3 Burger and the TurtleBot3 Waffle Pi have a low distance threshold for hardware protection. 
Consequently, the robots can form close groups as shown in Figs.~\ref{fig:attraction_burger} and~\ref{fig:attraction_waffle}. 
By contrast, the Jackal has a rather high distance threshold for hardware protection due to a high minimum detection range of its LiDAR. 
Hence, the robots do not group as closely as the TurtleBot3 robot platforms. 

\begin{figure}
    \centering
   \subfloat[Initial robot positions \label{fig:wp_initial}]{
    \includegraphics[width=0.48\linewidth]{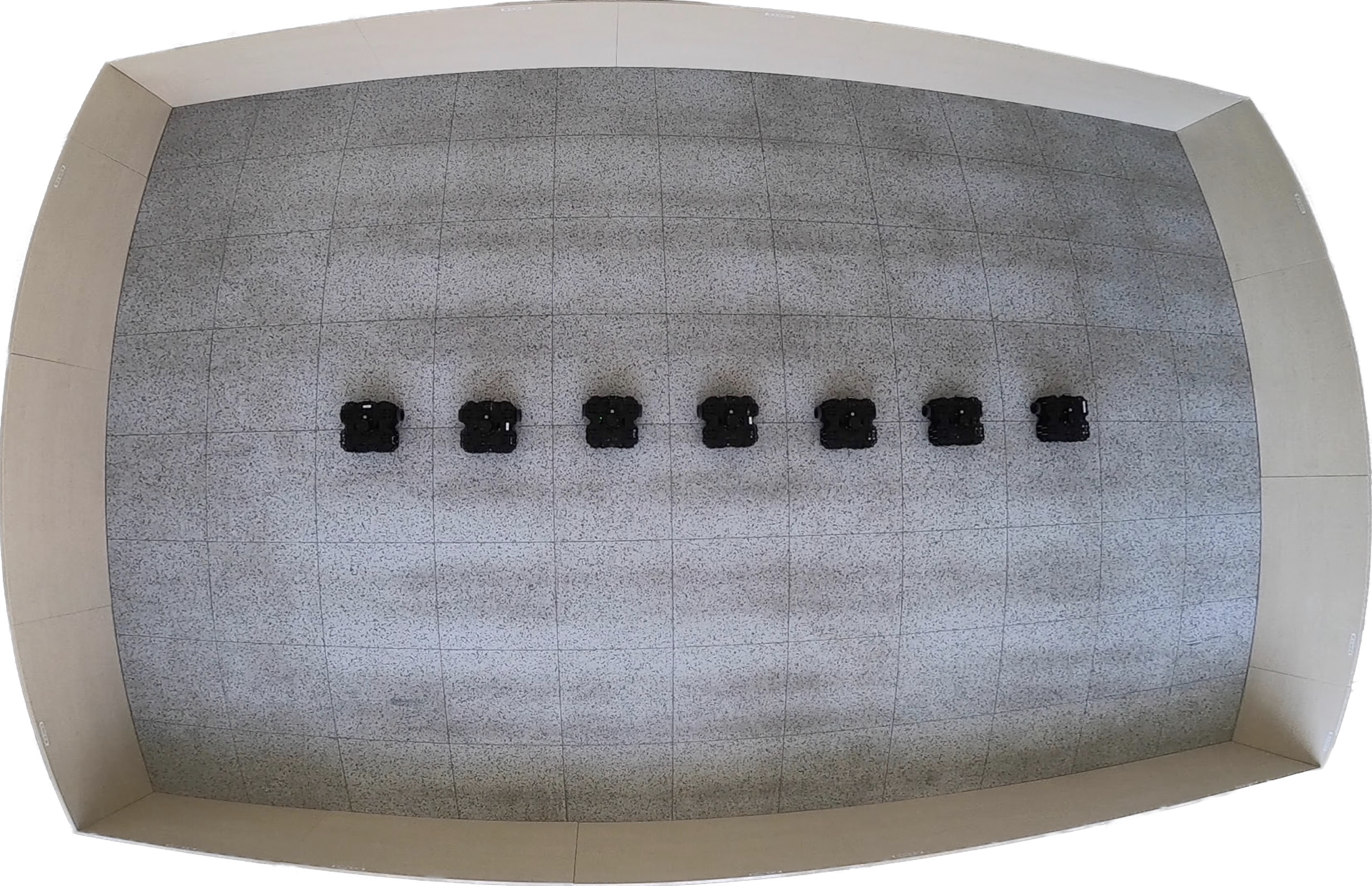}} 
   \subfloat[Aggregated swarm]{
    \includegraphics[width=0.465\linewidth]{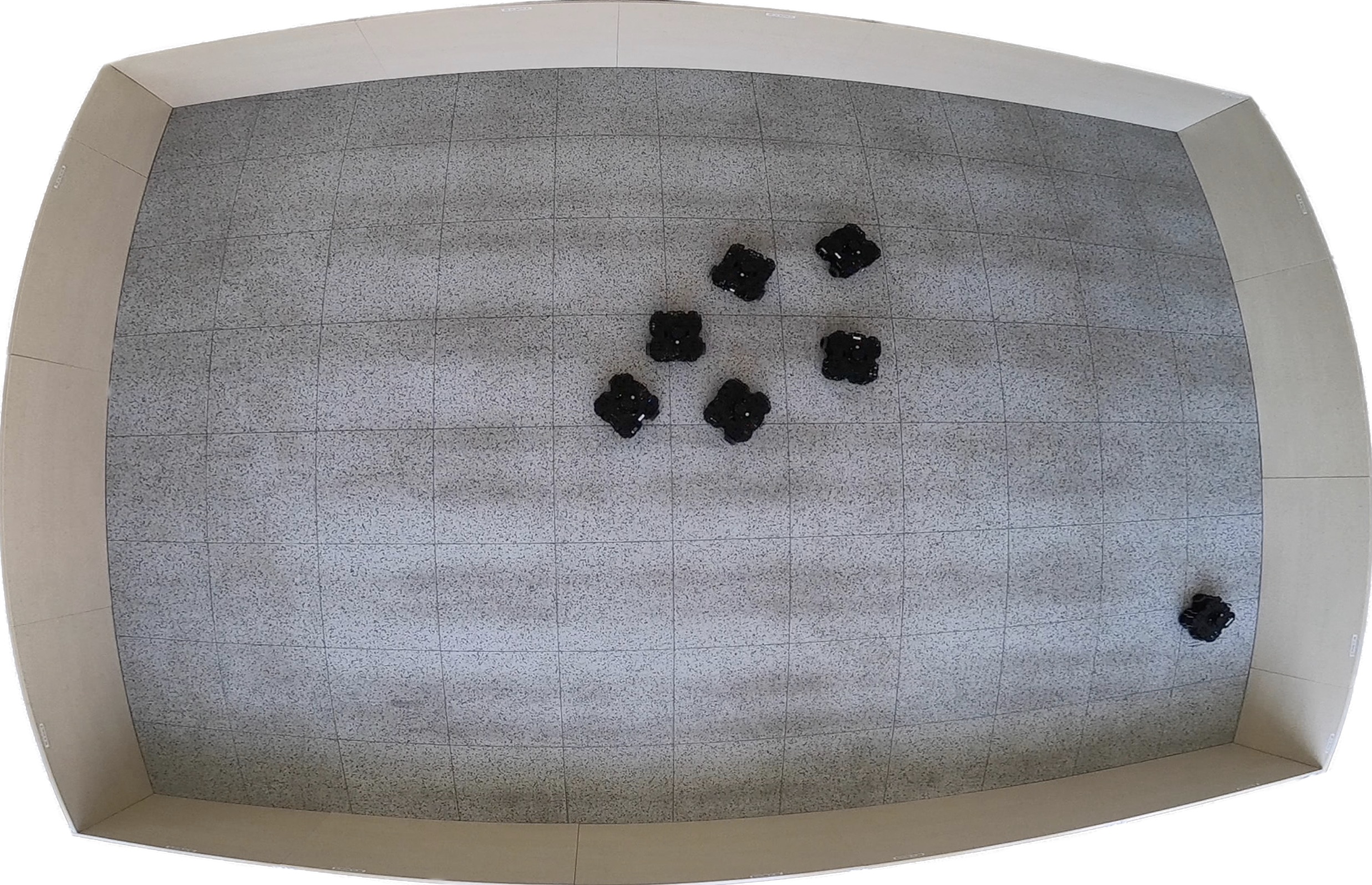}} 
\caption{Experiment 1 in hardware: a~swarm of seven TurtleBot3 Waffle Pis executing the attraction pattern. (a)~Top view of initial robot positions in a $4.8~\textrm{m}\times 6.6~\textrm{m}$ arena and (b)~aggregated robots. Six robots aggregate during pattern execution while the rightmost robot lost the group and  is attracted to the wall. \label{fig:wp_attraction} The floor tiles have a size of $0.6~\textrm{m}\times 0.6~\textrm{m}$. 
}
\end{figure}

Next, we run the attraction pattern in hardware on a swarm of seven real TurtleBot3 Waffle Pis in a $6.6~\textrm{m}\times 4.8~\textrm{m}$~empty arena that is bounded by walls. 
We set the attraction range from $0.12~\textrm{m}$ to $0.8~\textrm{m}$ to avoid that robots are mostly attracted to the arena walls due to the smaller arena. 
Fig.~\ref{fig:wp_attraction} shows the initial robot positions and the aggregated swarm. 
We find that six robots form one group that repeatedly splits and merges over time (see supplementary video). 
Thereby, two groups with three robots each form already after approx.~$5~\textrm{s}$ and merge the first time after around $60~\textrm{s}$ of execution time. 
One robot moved away from the group and is guided towards the arena's walls by the attractive potential field as robots and other obstacles cannot be differentiated based on the LiDAR's sensor data only. 
In summary, our \textsc{ROS2swarm} package proved to be effective in testing and running swarm behaviors on different robot platforms in simulations and on real robot hardware. 

\subsection{Experiment 2: Discussed Dispersion}\label{sec:cp}

\begin{figure}[tb]
    \centering
   \includegraphics[width=0.9\linewidth]{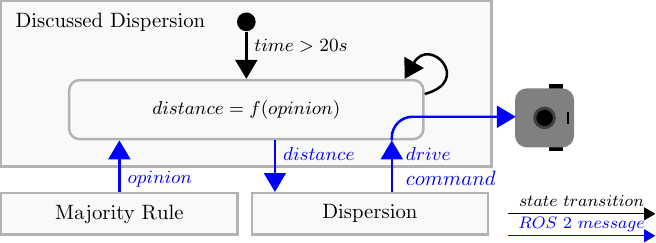}
\caption{Architecture of the \textit{discussed dispersion} pattern. 
For $20~\textrm{s}$, swarm members execute only the majority rule pattern to determine their majority opinion. This opinion is mapped to a distance by the discussed dispersion pattern using a mapping function~$f$ which is set as the minimum distance during dispersion.  
The dispersion pattern determines the robot's drive commands based on a repulsive potential field approach. 
Rectangular boxes give patterns and rounded boxes give states.
\label{fig:dd}
}
\end{figure}

In the next hardware experiment, we showcase how basic patterns can be used to form combined patterns to create more complex behaviors. 
This shows the modularity and reusability of the available patterns included in \textsc{ROS2swarm}'s swarm behavior library. 
As an example, we combine the basic majority rule voting pattern and the basic dispersion movement pattern and form a new pattern that we call \textit{discussed dispersion}. 
The robots are supposed to disperse while maintaining a robot-to-robot distance that they collectively choose themselves.
The minimum specifiable robot-to-robot distance equals the distance threshold triggering hardware protection as both dispersion and hardware protection layer use the same repulsive potential field approach.  
As visualized in Fig.~\ref{fig:dd}, robots execute only the majority rule pattern in the first $20~\textrm{s}$ staying on their initial positions (as previously shown in Fig.~\ref{fig:wp_initial}).  
After this initial decision phase, both the majority rule pattern and the dispersion pattern are executed in parallel. 
The swarm may still change its opinion, for example, when several new robots are added.  
The opinion determined by the majority rule pattern is mapped to a distance value using a mapping function~$f$. 
This is set as the minimum distance that robots maintain from all obstacles including other robots when executing the dispersion pattern. 

\begin{figure}
    \centering
    \subfloat[Initial/final opinions based on the majority rule]{\includegraphics[width=0.34\linewidth]{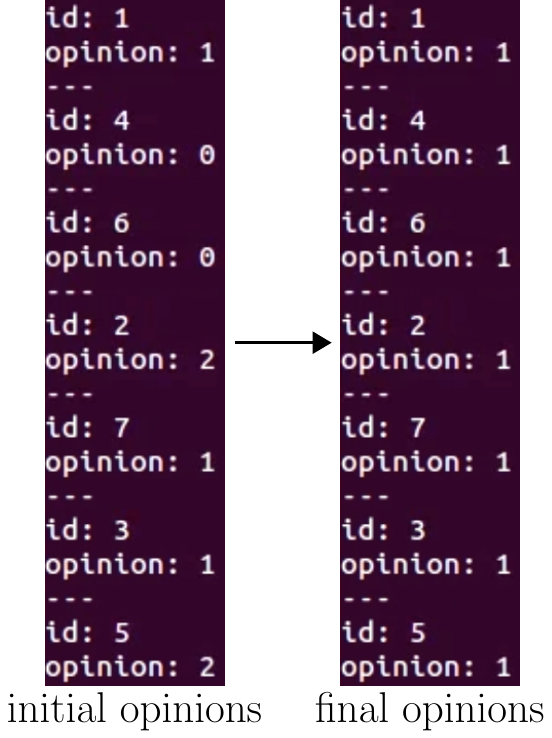}}
    \subfloat[Dispersed swarm at agreed on distance\label{fig:dd_final}]{
     \includegraphics[width=0.6\linewidth]{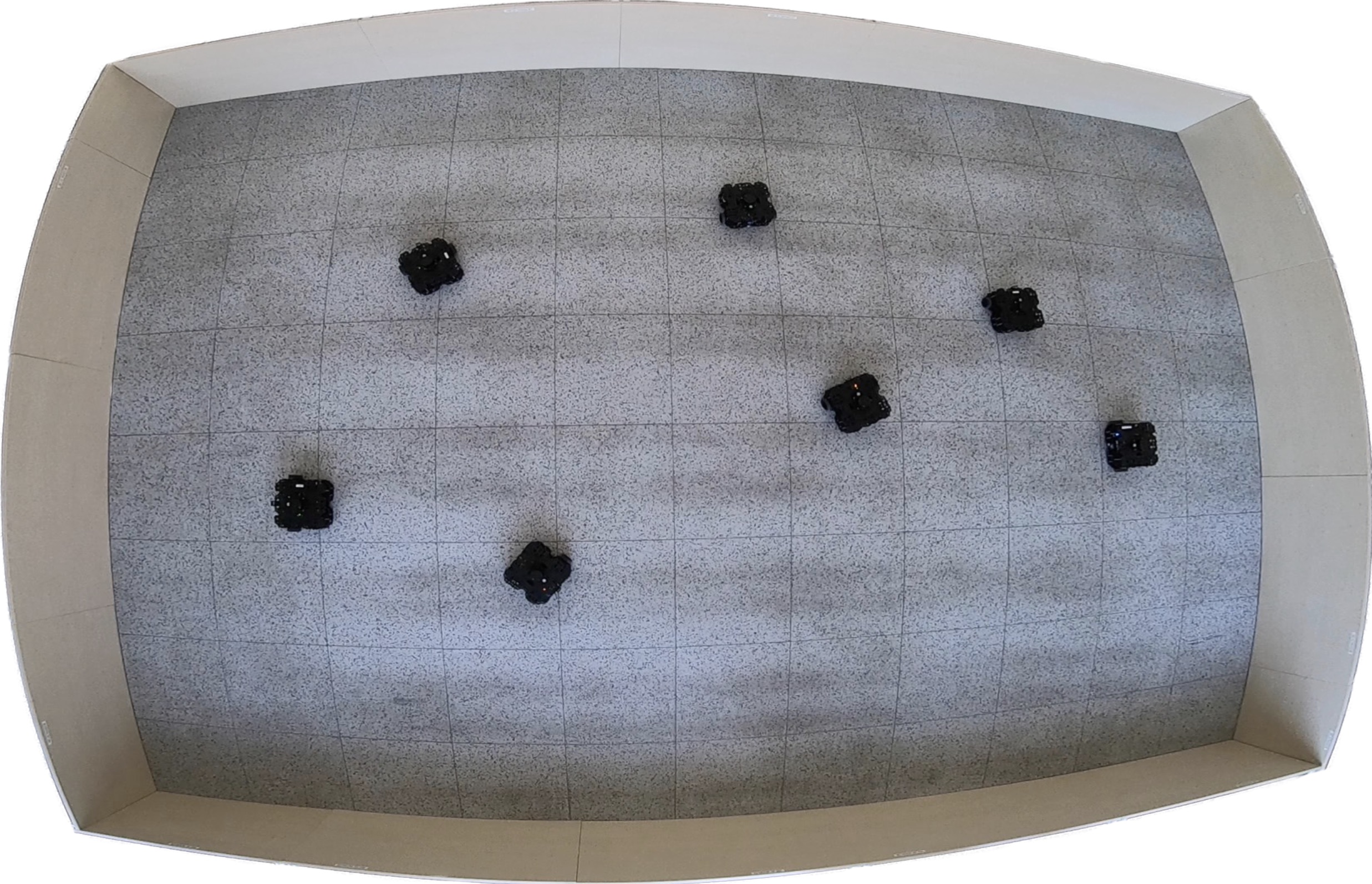}}
\caption{Experiment 2 in hardware: a swarm of seven~TurtleBot3 Waffle Pis executing the \textit{discussed dispersion} pattern that combines two basic patterns. The robots keep the start configuration as shown in Fig.~\ref{fig:wp_initial} for the first $20~\textrm{s}$ executing only the majority rule pattern. In this case here, robots decide for opinion~$1$ that is mapped to a minimal distance between robots of~$1.0~\textrm{m}$. Afterwards, robots disperse in the arena keeping this collectively decided distance (or more) to each other and the arena walls. \label{fig:dd_gazebo}}
\end{figure}

We execute the combined pattern on a swarm of seven real TurtleBot3 Waffle Pis, see Fig.~\ref{fig:dd_gazebo}. 
The opinion of each robot is randomly initialized to $r \in \{0, 1, 2\}$ and mapped to a distance of $f(0)=0.6~\textrm{m}$, $f(1)=1.0~\textrm{m}$, or $f(2)=1.4~\textrm{m}$, respectively. 
In the visualized run, robots collectively choose opinion~$1$ (distance of~$1.0~\textrm{m}$). 
As shown in Fig.~\ref{fig:dd_gazebo}, the swarm distributes in space until each robot is at least $1.0~\textrm{m}$ away from all obstacles, that is, other robots and walls. In summary, we have shown how the modularity of our swarm behavior library can easily be leveraged to derive new and more complex swarm behaviors.

\section{Conclusion} \label{sec:conclusion} 

Our \textsc{ROS2swarm} ROS~2 package provides a library of ready-to-use swarm behaviors for collective motion and collective decision-making applications. 
We showed that the package can be extended easily and that all patterns are reusable and modular. 
This allows to run combined patterns for more complex applications as illustrated in our Experiment~2 (cf. Sec.~\ref{sec:cp}). 
To date, three ground mobile robots are supported out of the box by \textsc{ROS2swarm}, namely  TurtleBot3 Burger,  TurtleBot3 Waffle Pi, and  Jackal. 
Support for further robot platforms running ROS or ROS~2 can be integrated with minimal effort. 
\textsc{ROS2swarm} unites the advantages of swarm robotics, namely adaptability, robustness, and scalability, and of ROS~2, that is, platform independence and modularity.   
In total, our package is versatile and enables a quick and easy setup of swarm robotics applications. 

While the package in its current form already offers the most important concepts for swarm applications, we want to add several enhancements in the future. 
First, all robots must currently be connected to a common network for the execution of voting patterns. 
So far, we use a central WiFi network with a single access point for this purpose and thus have a potential single point of failure. 
In future, we will integrate IR communication~\cite{Arvin2009} or switch to an ad-hoc WiFi network that allows to determine a robot's neighbors based on the decreasing WiFi signal strength over distance. 
Also, our hardware protection layer does not guarantee the prevention of collisions using the current repulsive potential field approach. 
We plan to investigate more reliable approaches, such as decentralized safety barrier certificates~\cite{wang2017}. 
Furthermore, we continue to extend the library with more collective behaviors and, in particular, with more complex behaviors, such as pattern formation. 
We will integrate tests into the package for quantitatively evaluating the effectiveness of implemented swarm behaviors, for example, by computing the mean-distance-to-centroid in aggregation.
A~longer-term plan is to integrate support for more robot platforms that may need different drive commands (e.g., quadcopter swarms~\cite{barcis2020sandsbots,steup2020generic}) than the currently included platforms and to make the package independent of a LiDAR by supporting different sensor setups for obstacle detection, such as multiple IR sensors and cameras. 
We plan to release the package to the public ROS~2 buildfarm. 
We hope to form a developer community that keeps extending and maintaining this package for the profit of a growing swarm robotics research and teaching community.

\section*{Acknowledgments}

The authors thank Steffen Fleischmann, Vincent Jansen, and Daniel Tidde for their contributions to the implementation of \textsc{ROS2swarm}, Yuanchen Yuan for proofreading the paper, and Lucas Kaiser for support in image editing. 

\bibliographystyle{IEEEtran}  
\bibliography{references}

\begin{thebibliography}{10}
\providecommand{\url}[1]{#1}
\csname url@rmstyle\endcsname
\providecommand{\newblock}{\relax}
\providecommand{\bibinfo}[2]{#2}
\providecommand\BIBentrySTDinterwordspacing{\spaceskip=0pt\relax}
\providecommand\BIBentryALTinterwordstretchfactor{4}
\providecommand\BIBentryALTinterwordspacing{\spaceskip=\fontdimen2\font plus
\BIBentryALTinterwordstretchfactor\fontdimen3\font minus \fontdimen4\font\relax}
\providecommand\BIBforeignlanguage[2]{{%
\expandafter\ifx\csname l@#1\endcsname\relax
\typeout{** WARNING: IEEEtran.bst: No hyphenation pattern has been}%
\typeout{** loaded for the language `#1'. Using the pattern for}%
\typeout{** the default language instead.}%
\else
\language=\csname l@#1\endcsname
\fi
#2}}

\bibitem{Hamann2018}
H.~Hamann, \emph{Swarm Robotics: A Formal Approach}.\hskip 1em plus 0.5em minus 0.4em\relax Cham: Springer International Publishing, 2018.

\bibitem{nesnas2006claraty}
I.~A.~D. Nesnas, R.~Simmons, D.~Gaines, C.~Kunz, A.~Diaz-Calderon, T.~Estlin, R.~Madison, J.~Guineau, M.~McHenry, I.-H. Shu, and D.~Apfelbaum, ``{CLARAty}: Challenges and steps toward reusable robotic software,'' \emph{International Journal of Advanced Robotic Systems}, vol.~3, no.~1, p.~5, 2006.

\bibitem{pinciroli2016buzz}
C.~Pinciroli and G.~Beltrame, ``Buzz: a programming language for robot swarms,'' \emph{IEEE Software}, vol.~33, no.~4, pp. 97--100, 2016.

\bibitem{quigley2009ros}
M.~Quigley, J.~Faust, T.~Foote, and J.~Leibs, ``{ROS}: an open-source {R}obot {O}perating {S}ystem,'' in \emph{ICRA workshop on open source software}, vol.~3, no. 3.2.\hskip 1em plus 0.5em minus 0.4em\relax Kobe, Japan, 2009, p.~5.

\bibitem{dorigo21}
M.~Dorigo, G.~Theraulaz, and V.~Trianni, ``Swarm robotics: Past, present, and future [point of view],'' \emph{Proceedings of the IEEE}, vol. 109, no.~7, pp. 1152--1165, 2021.

\bibitem{Maruyama2016}
Y.~{Maruyama}, S.~{Kato}, and T.~{Azumi}, ``Exploring the performance of {ROS}2,'' in \emph{2016 International Conference on Embedded Software (EMSOFT)}, 2016, pp. 1--10.

\bibitem{barcis2019}
A.~{Barciś}, M.~{Barciś}, and C.~{Bettstetter}, ``Robots that {S}ync and {S}warm: A proof of concept in {ROS} 2,'' in \emph{2019 International Symposium on Multi-Robot and Multi-Agent Systems (MRS)}, 2019, pp. 98--104.

\bibitem{barcis2020sandsbots}
A.~Barci{\'s} and C.~Bettstetter, ``Sandsbots: Robots that sync and swarm,'' \emph{IEEE Access}, vol.~8, pp. 218\,752--218\,764, 2020.

\bibitem{Testa2021}
A.~{Testa}, A.~{Camisa}, and G.~{Notarstefano}, ``Choi{R}bot: A {ROS}~2 toolbox for cooperative robotics,'' \emph{IEEE Robotics and Automation Letters}, vol.~6, no.~2, pp. 2714--2720, 2021.

\bibitem{queralta2021towards}
J.~P. Queralta, Y.~Xianjia, L.~Qingqing, and T.~Westerlund, ``Towards large-scale scalable {MAV} swarms with {ROS}2 and {UWB}-based situated communication.''

\bibitem{deWolf2006}
T.~De~Wolf and T.~Holvoet, ``Design patterns for decentralised coordination in self-organising emergent systems,'' in \emph{Proceedings of the 4th International Conference on Engineering Self-Organising Systems}, ser. ESOA'06.\hskip 1em plus 0.5em minus 0.4em\relax Berlin, Heidelberg: Springer-Verlag, 2006, p. 28–49.

\bibitem{fernandez-marquez_description_2013}
J.~L. Fernandez-Marquez, G.~Di~Marzo~Serugendo, S.~Montagna, M.~Viroli, and J.~L. Arcos, ``\BIBforeignlanguage{en}{Description and composition of bio-inspired design patterns: a complete overview},'' \emph{\BIBforeignlanguage{en}{Natural Computing}}, vol.~12, no.~1, pp. 43--67, Mar. 2013.

\bibitem{pitonakova2018}
L.~Pitonakova, R.~Crowder, and S.~Bullock, ``Information exchange design patterns for robot swarm foraging and their application in robot control algorithms,'' \emph{Frontiers in Robotics and AI}, vol.~5, p.~47, 2018.

\bibitem{stonge2020}
D.~St-Onge, V.~S. Varadharajan, I.~Švogor, and G.~Beltrame, ``From design to deployment: Decentralized coordination of heterogeneous robotic teams,'' \emph{Frontiers in Robotics and AI}, vol.~7, p.~51, 2020.

\bibitem{micros_swarm_framework}
\BIBentryALTinterwordspacing
C.~Xuefeng, C.~Zhongxuan, W.~Yanzhen, and Y.~Xiaodong, ``micros\_swarm\_framework - {ROS} {Wiki},'' 2019. [Online]. Available: \url{https://wiki.ros.org/micros_swarm_framework}
\BIBentrySTDinterwordspacing

\bibitem{morris2018full}
K.~Morris, G.~Arpino, S.~Nagavalli, and K.~Sycara, ``Full stack swarm architecture,'' \emph{RISS Working Papers Journal}, 2018.

\bibitem{hrabia2018}
C.-E. Hrabia, T.~K. Kaiser, and S.~Albayrak, ``Combining self-organisation with decision-making and planning,'' in \emph{Multi-Agent Systems and Agreement Technologies}, F.~Belardinelli and E.~Argente, Eds.\hskip 1em plus 0.5em minus 0.4em\relax Cham: Springer International Publishing, 2018, pp. 385--399.

\bibitem{KalempaTeixeiraOliveiraFabro2018}
V.~C. Kalempa, M.~A.~S. Teixeira, A.~S. de~Oliveira, and J.~A. Fabro, ``Intelligent dynamic formation of the multi-robot systems to cleaning tasks in unstructured environments and with a single perception system,'' in \emph{2018 Latin American Robotic Symposium, 2018 Brazilian Symposium on Robotics (SBR) and 2018 Workshop on Robotics in Education (WRE)}, Nov 2018, pp. 71--76.

\bibitem{swarm_robot_ros_sim}
\BIBentryALTinterwordspacing
Y.~Liu, A.~Ali, and G.~Dare, ``swarm\_robot\_ros\_sim,'' 2020. [Online]. Available: \url{https://github.com/yangliu28/swarm_robot_ros_sim}
\BIBentrySTDinterwordspacing

\bibitem{gazebo_webpage}
\BIBentryALTinterwordspacing
Gazebo, ``Gazebo,'' 2020. [Online]. Available: \url{http://gazebosim.org/}
\BIBentrySTDinterwordspacing

\bibitem{moeslinger2011}
C.~Moeslinger, T.~Schmickl, and K.~Crailsheim, ``A minimalist flocking algorithm for swarm robots,'' in \emph{Advances in Artificial Life. Darwin Meets von Neumann}, G.~Kampis, I.~Karsai, and E.~Szathm{\'a}ry, Eds.\hskip 1em plus 0.5em minus 0.4em\relax Berlin, Heidelberg: Springer Berlin Heidelberg, 2011, pp. 375--382.

\bibitem{valentini16a}
G.~Valentini, E.~Ferrante, H.~Hamann, and M.~Dorigo, ``Collective decision with 100 {K}ilobots: Speed vs accuracy in binary discrimination problems,'' \emph{Journal of Autonomous Agents and Multi-Agent Systems}, vol.~30, no.~3, pp. 553--580, 2016.

\bibitem{valentini2017best}
G.~Valentini, E.~Ferrante, and M.~Dorigo, ``The best-of-n problem in robot swarms: Formalization, state of the art, and novel perspectives,'' \emph{Frontiers in Robotics and AI}, vol.~4, p.~9, 2017.

\bibitem{Thomas2015}
\BIBentryALTinterwordspacing
D.~Thomas, E.~Fernandez, and W.~Woodall, ``State of {ROS~2} - demos and the technology behind,'' in \emph{{ROSCon} Hamburg 2015}.\hskip 1em plus 0.5em minus 0.4em\relax Open Robotics, September 2015. [Online]. Available: \url{https://doi.org/10.36288/ROSCon2015-900743}
\BIBentrySTDinterwordspacing

\bibitem{Arvin2009}
F.~Arvin, K.~Samsudin, and A.~R. Ramli, ``A short-range infrared communication for swarm mobile robots,'' in \emph{2009 International Conference on Signal Processing Systems}, 2009, pp. 454--458.

\bibitem{wang2017}
L.~Wang, A.~D. Ames, and M.~Egerstedt, ``Safety barrier certificates for collisions-free multirobot systems,'' \emph{IEEE Transactions on Robotics}, vol.~33, no.~3, pp. 661--674, 2017.

\bibitem{steup2020generic}
C.~Steup, S.~Parlow, S.~Mai, and S.~Mostaghim, ``Generic component-based mission-centric energy model for micro-scale unmanned aerial vehicles,'' \emph{Drones}, vol.~4, no.~4, p.~63, 2020.

\end{thebibliography}

\end{document}